%% file: asc.tex
\documentclass{article}
\newcommand{\modelcode}{ASC\xspace}
\newcommand{\modelname}{Adaptive Superpixel Coding\xspace}
\newcommand{\modelasplayer}{Adaptive Superpixel Layer\xspace}

\newcommand{\simclr}{\textsc{SimCLR}\xspace}

\newcommand{\byol}{\textsc{BYOL}\xspace}

\newcommand{\mae}{\textsc{MAE}\xspace}
\newcommand{\clip}{\textsc{CLIP}\xspace}
\newcommand{\dino}{\textsc{DINO}\xspace}
\newcommand{\toMe}{\textsc{ToMe}\xspace}
\newcommand{\piToMe}{\textsc{PiToMe}\xspace}
\newcommand{\dinoVtwo}{\textsc{DINOv2}\xspace}

\newcommand{\supervisedbaseline}{Supervised-IN \xspace}

\usepackage{ijcai25}

\usepackage[utf8]{inputenc}       
\usepackage[T1]{fontenc}          
\usepackage{times}                
\usepackage{soul}                 
\usepackage{url}                  
\usepackage[hidelinks]{hyperref} 
\usepackage[small]{caption}      
\usepackage{subcaption}          
\usepackage{graphicx}            
\usepackage{booktabs}            
\usepackage{multirow}            
\usepackage{colortbl}            
\usepackage[dvipsnames]{xcolor}  
\usepackage{lscape}              
\usepackage{nicefrac}            
\usepackage{microtype}           
\usepackage{xspace}              
\usepackage{amsmath, amssymb, amsfonts, mathrsfs} 
\usepackage{amsthm}              
\usepackage{algorithm}
\usepackage{algpseudocode}
\usepackage[switch]{lineno}      
\usepackage{placeins} 

\definecolor{pearDark}{RGB}{153, 204, 0} 

\title{Representation Learning with Adaptive Superpixel Coding}

\author{
Mahmoud Khalil$^1$\and
Ahmad Khalil$^1$\and
Alioune Ngom$^1$\\
\affiliations
$^1$School of Computer Science, University of Windsor, Windsor, ON, Canada\\
\emails
\{khalil12, khalil71, angom\}@uwindsor.ca
}

\begin{document}

\maketitle

\begin{abstract}
    \vspace{-0.7em}
    \input{sections/0_abstract}  
\end{abstract}
\vspace{-0.7em}

\vspace{-0.7em}
\input{sections/1_introduction}
\input{sections/2_related_work}
\input{sections/3_method}

\input{sections/4_experiments}
\input{sections/5_model_ablations}

\input{sections/6_conclusion}

\bibliographystyle{plain}  
\bibliography{asc}

\input{sections/z_appendix}

\end{document}

%% file: sections/0_abstract.tex
Deep learning vision models are typically tailored for specific modalities and often rely on domain-specific assumptions, such as the grid structures used by nearly all existing vision models.  
In this work, we propose a self-supervised model based on Transformers, which we call \modelname (\modelcode).  
The key insight of our model is to overcome the limitations of traditional Vision Transformers, which depend on fixed-size and non-adaptive patch partitioning. Instead, \modelcode employs adaptive superpixel layers that dynamically adjust to the underlying image content.  

We analyze key properties of the approach that make it effective, and find that our method outperforms widely-used alternatives on standard image downstream task benchmarks.  

%% file: sections/1_introduction.tex
\section{Introduction}
\label{sec:introduction}

Deep learning has revolutionized computer vision by enabling models to learn meaningful representations directly from unstructured sensory data, yielding significant performance gains across a wide range of tasks~\cite{bommasani2021opportunities}. In particular, self-supervised learning (SSL) has emerged as a powerful paradigm for learning visual representations without human annotations, achieving remarkable results in image classification, segmentation, and detection.

However, most SSL methods rely on fixed grid-based representations to encode image structure. Convolutional Neural Networks (CNNs)~\cite{he2016deep,sandler2018mobilenetv2} treat the image as a regular grid and extract local features using sliding windows. Vision Transformers (ViTs)~\cite{dosovitskiy2020image}, though more flexible in modeling long-range dependencies, It begins by decomposing an image into non-overlapping patches arranged in a uniform grid.
These grid structures lead to an inherent entanglement between the representation structure and image structure, constraining the model's ability to adapt to variation in object shape, scale, or layout.

This entanglement between the representation structure and the image structure is especially limiting in downstream scene understanding tasks, where object-level reasoning, deformation invariance, and flexible spatial grouping are often essential. Many of these tasks require the model to localize, compare, and track object parts across varying spatial arrangements—capabilities poorly supported by rigid, grid-aligned tokens.

Superpixels offer a natural alternative. Originating as a form of image over-segmentation~\cite{ren2003learning}, superpixels group pixels into visually and semantically coherent regions. These regions serve as low- to mid-level primitives that align well with object boundaries and exhibit strong inductive priors for efficient computation. They have been widely used in vision tasks including object detection~\cite{shu2013improving,yan2015object}, semantic segmentation~\cite{gould2008multi,sharma2014recursive}, saliency estimation~\cite{he2015supercnn}, and optical flow~\cite{hu2016highly}, among others.

Yet, despite their utility, superpixels remain underutilized in transformer-based models, with only a few exceptions~\cite{mei2024spformer,zhu2023superpixel}. One key reason is that transformers assume a fixed token layout, typically regular patches. However, superpixels are inherently irregular and content-adaptive, which clashes with the design of existing architectures that assume spatial consistency across layers.

In this work, we revisit the idea of superpixels and propose a token-level grouping mechanism for transformers. Rather than enforcing grid-based token structures, we introduce a novel transformer-compatible layer, \textit{\modelname (\modelcode)}, that adaptively merges tokens into semantically coherent regions. This layer computes pairwise token similarities, forms a weighted graph, and extracts connected components via a differentiable thresholding and grouping operation. The resulting components serve as \textit{superpixels} that replace fixed patches in subsequent transformer layers.

Importantly, we demonstrate that this representation is especially effective in a self-supervised setting. Using contrastive training based on frame-level similarity across videos. The resulting representations are more aligned with object structure and transfer well to diverse downstream tasks.

\textbf{Our contributions are summarized as follows:}
\begin{itemize}
   \item We introduce \textit{\modelname (\modelcode)}, a novel transformer-compatible layer that integrates an adaptive superpixel mechanism, enabling the decoupling of the image's \textit{grid-based} structure from its representation structure.
   \item We demonstrate that the learned representation transfers effectively to a wide range of downstream vision tasks, including image classification, object detection, and semantic segmentation. Our method outperforms strong self-supervised baselines such as \byol~\cite{grill2020bootstrap} and \dino~\cite{caron2021emerging}.
\end{itemize}

%% file: sections/2_related_work.tex
\section{Related Work}
\label{sec:related-work}
\paragraph{Grid-based Representation.}Deep learning methods for visual tasks \cite{krizhevsky2012imagenet,koonce2021resnet,chen2020simple,dosovitskiy2020image} have garnered significant attention in the machine learning community in recent years. Early work, such as the use of CNN feature maps \cite{krizhevsky2012imagenet}, laid the foundation, followed by several efforts to integrate CNN-like architectures with self-attention mechanisms \cite{wang2018non,carion2020end}, and even replace convolutions altogether \cite{ramachandran2019stand,wang2020axial}. Later innovations, such as the Vision Transformer (ViT) \cite{dosovitskiy2020image}, represent a natural evolution of this trend. Most of these approaches operate on a grid of image features ranging from patches or CNN feature maps to the final output layer. Although this grid-based representation has proven highly effective across a broad range of tasks, it inherently ties the learned features to fixed spatial 2D locations in image space.

\paragraph{Off-the-Grid Representation.}A key method for decoupling model representations from the image grid is through the use of \textit{cross-attention}, where one set of tokens is updated based on the values of another set. Related approaches, such as \cite{locatello2020object,kipf2021conditional}, extend this idea by leveraging slot-based attention mechanisms to model object-centric representations. The cross-attention approach has been particularly effective in object-centric tasks such as detection \cite{carion2020end,zhu2020deformable}, tracking \cite{kipf2021conditional,meinhardt2022trackformer}, and instance segmentation \cite{cheng2022masked,kirillov2023segment}. 
GroupViT~\cite{xu2022groupvit} is closely related to our approach, as it employs a large number of group tokens that are updated via cross-attention from visual tokens. In this framework, each group token serves as a compact and learnable representation that captures both local and global contextual information from the image. These tokens interact with image tokens through a series of cross-attention operations.

Inspired by GroupViT, our model introduces a novel layer into the Vision Transformer (ViT) architecture. However, in contrast to GroupViT, our method does not rely on explicit object queries or grouping tokens. Instead, we perform adaptive thresholding on similarity scores to modulate the resulting representations, allowing for more flexible and context-sensitive feature aggregation.

\paragraph{Combining Tokens.}
Our work is closely related to recent efforts aimed at improving the efficiency of Vision Transformers by reducing the number of tokens processed per layer. Notably, Token Merging (ToMe)~\cite{bolya2022token} proposes a deterministic token coalescing mechanism based on token similarity, enabling faster inference without retraining. Other methods, such as AdaViT~\cite{meng2022adavit} and DynamicViT~\cite{rao2021dynamicvit}, introduce adaptive token pruning or merging strategies based on confidence scores or learned importance measures.

While inspired by this broader class of approaches, our method differs in both \textit{motivation} and \textit{mechanism}.Instead of focusing on token efficiency, our goal is to induce \textit{object-centric representations} by constructing a token affinity graph and identifying \textit{connected components} to form semantically coherent groups.

In contrast to ToMe, which merges token pairs based on a greedy bipartite matching algorithm, our method constructs a \textit{soft adjacency matrix} using a learnable gating mechanism and discovers groups via graph traversal (DFS). This enables us to form compositional groupings of similar tokens, rather than pairwise merges. Our formulation introduces two key distinctions: (1) a \textit{differentiable connectivity threshold} jointly learned with the model, and (2) a \textit{graph-theoretic} interpretation of token affinity, which structures the merging process around spatial and semantic coherence.

\paragraph{Superpixels.}
Recent methods such as SpFormer~\cite{mei2024spformer} and Superpixel Transformer~\cite{zhu2023superpixel} incorporate superpixel-inspired structures into Transformer architectures to promote inductive biases that align with object boundaries. These methods \textit{learn} to group pixels based on high-level semantic features through task-driven supervision (classification or segmentation objectives), rather than optimizing for low-level features of the pixel space. Although they incorporate a locality prior that biases adjacent pixels to cluster together, superpixel formation is ultimately guided by discriminative task performance. 
In contrast, our approach explicitly constructs a token affinity graph based on pairwise token similarity and performs grouping via a soft gating mechanism followed by graph traversal.

%% file: sections/3_method.tex
\input{figures/text/1_pipline_figure}
\section{Method}
\label{sec:method}
In this section, we introduce \modelname (Fig.~\ref{fig:concept}; Section ~\ref{sec:method-adaptive-superpixel-layer}) and demonstrate how
it can be integrated into an architecture for self-supervised contrastive learning (Fig.~\ref{fig:pipeline}; Section ~\ref{sec:method-self-sup}).

\subsection{Overall Architecture}
\label{sec:method-self-sup}

An overview of the \modelname{} Transformer architecture is presented in (Fig.~\ref{fig:pipeline}).

Inspired by Video Frame-level Similarity (VFS)~\cite{xu2021rethinking}, we begin by sampling two random frames, \( f_i \) and \( f_j \), from a video clip, where each frame is represented as an RGB image \( f \in \mathbb{R}^{H \times W \times 3} \). Each frame is uniformly partitioned into \( N \) non-overlapping patches, which serve as basic visual tokens.

For each patch, the raw RGB values are flattened into a vector of dimension \( 4 \times 4 \times 3 = 48 \) (assuming a patch size of \( 4 \times 4 \)), and subsequently mapped into a latent embedding space of dimension \( C \) via a linear projection layer. This process yields token sets \( \{\mathbf{z}^{(i)}_t\}_{t=1}^T \) and \( \{\mathbf{z}^{(j)}_t\}_{t=1}^T \) for frames \( f_i \) and \( f_j \), respectively, where superscripts indicate the source frame and subscripts index the patch.

The resulting patch embeddings are then processed by a series of Transformer blocks, equipped with an \modelasplayer~\ref{sec:method-adaptive-superpixel-layer}. These token sequences, derived independently from frames \( f_i \) and \( f_j \), are passed through two branches of a Siamese architecture: a predictor encoder \( P \) and a target encoder \( Q \). Each encoder operates on the full token sequence, producing contextualized embeddings that are subsequently normalized using the L2 norm:

\begin{equation}
\mathbf{Z_i} = \frac{P\left(\{\mathbf{z}^{(i)}_t\}_{t=1}^T\right)}{\left\|P\left(\{\mathbf{z}^{(i)}_t\}_{t=1}^T\right)\right\|_2}, \quad 
\mathbf{Z_j} = \frac{Q\left(\{\mathbf{z}^{(j)}_t\}_{t=1}^T\right)}{\left\|Q\left(\{\mathbf{z}^{(j)}_t\}_{t=1}^T\right)\right\|_2}
\label{eq:normalized_proj}
\end{equation}
where \( \mathbf{Z_i} \) and \( \mathbf{Z_j} \) denote the normalized embeddings of frames \( f_i \) and \( f_j \), respectively, produced by the predictor encoder \( P \) and the target encoder \( Q \). If these embeddings capture similar content, contrastive learning encourages their alignment while pushing apart representations from unrelated samples~\cite{wu2018unsupervised}.

\subsection{Adaptive Superpixel Block}
\label{sec:method-adaptive-superpixel-layer}

As illustrated in Fig.~\ref{fig:concept}, the Adaptive Superpixel Block consists of a standard self-attention layer followed by \modelasplayer{}.

\input{figures/text/0_concept_figure}

Given contextualized token embeddings produced by self-attention, we compute a superpixel-like representation by interpreting pairwise similarities as edges in a graph. Rather than using a full attention distribution, we adopt the dot product to compute affinities:

\begin{equation}
    S = \mathbf{Z} \mathbf{Z}^\top, \qquad \text{where } \mathbf{Z} \in \mathbb{R}^{N \times d}
\label{eq:dot_product_similarity}
\end{equation}

Each entry \( S_{ij} \) quantifies the similarity between token embeddings \( z_i \) and \( z_j \). Since \( S \) is symmetric, it can be interpreted as a weighted adjacency matrix of an undirected graph \( G = (V, E) \), where each token is a node, and the edge weights reflect the semantic affinity.

\paragraph{Theoretical Foundations and Assumptions.}

\begin{itemize}
    \item \textbf{Undirected Graph:} Since \( S^\top = S \), the graph \( G \) is undirected. That is, if \( (i, j) \in E \), then \( (j, i) \in E \) and \( s_{ij} = s_{ji} \).
    
    \item \textbf{Graph Model:} Let \( V \) denote the set of tokens, and let an edge \( (i, j) \in E \) exist if \( A_{ij} > 0 \), where
    \[
    A = \sigma(S - \theta),
    \]
     \( \sigma \) is the sigmoid function and \( \theta \) is a learnable threshold. The matrix \( A \in [0,1]^{N \times N} \) encodes soft connectivity and controls the sparsity of the graph.

    \item \textbf{Connectedness and Objecthood:} Let \( \sim \) be the binary relation over \( V \) such that \( i \sim j \) if and only if there exists a path in \( G \) connecting \( z_i \) and \( z_j \). The relation \( \sim \) is an equivalence relation (reflexive, symmetric, transitive), and partitions \( V \) into disjoint connected components. We formally define an \textbf{object} as such a connected component, consistent with Proposition~1 and~2.
\end{itemize}

\paragraph{\modelasplayer{} Mechanism.}

The goal of \modelasplayer{} is to construct object-centric embeddings by identifying and merging connected components in the similarity graph.

Let \( K = [k_1, \dots, k_N] \in \mathbb{R}^{N \times d_z} \) denote the key representations associated with tokens \( Z = [z_1, \dots, z_N] \). Define the similarity matrix as:

\begin{equation}
S_{ij} = k_i^\top k_j, \qquad A = \sigma(S - \theta),
\end{equation}

A graph \( G = (V, E) \) is constructed where \( z_i, z_j \in V \) are connected by an undirected edge \( (i, j) \in E \) if \( A_{ij} > 0 \). By construction, \( A \) is symmetric and thus defines an undirected graph.

\textbf{Proposition 1 (Object Membership).}  
Two tokens \( z_i \) and \( z_j \) belong to the same object if and only if there exists a path in \( G \) connecting them.

\textbf{Proposition 2 (Transitivity of Connectivity).}  
If \( z_i \sim z_j \) and \( z_j \sim z_k \), then \( z_i \sim z_k \). Hence, all three tokens belong to the same connected component, forming a single object.

\textit{The proof is in the supplementary material.}

\paragraph{Component Aggregation.}

The algorithmic implementation uses depth-first search (DFS) to enumerate connected components in \( G \), as outlined in Algorithm~\ref{alg:dfs}. Each connected component \( C_i \subseteq V \) defines a group of semantically coherent tokens.

\begin{algorithm}[H]
\scriptsize
\caption{Depth-First Search (DFS) for Connected Components}
\label{alg:dfs}
\begin{algorithmic}[1]
\Function{DFS}{node, $A$, vis, curr}
    \State vis[node] $\gets$ True
    \State append node to curr
    \For{nbr = 0 to $N{-}1$}
        \If{$A[\text{node}, \text{nbr}] > 0$ \textbf{and} $\neg$ vis[nbr]}
            \State \Call{DFS}{nbr, $A$, vis, curr}
        \EndIf
    \EndFor
\EndFunction

\Function{FindComponents}{$A$}
    \State vis $\gets$ array of size $N$ initialized to False
    \State comps $\gets$ empty list
    \For{node = 0 to $N{-}1$}
        \If{$\neg$ vis[node]}
            \State curr $\gets$ empty list
            \State \Call{DFS}{node, $A$, vis, curr}
            \State append curr to comps
        \EndIf
    \EndFor
    \State \Return comps
\EndFunction
\end{algorithmic}
\end{algorithm}

\paragraph{Feature Merging.}
Once connected components are identified, the embeddings within each component \( C_i \subseteq \{1, \dots, N\} \) are merged via mean pooling:

\begin{equation}
\mathbf{z}^{l + 1}_i = \frac{1}{|C_i|} \sum_{j \in C_i} \mathbf{z}_j^l
\label{eq:component_merge}
\end{equation}

This yields a compact set of embeddings \( \tilde{Z} \in \mathbb{R}^{\tilde{N} \times d_z} \), where \( \tilde{N} \leq N \), and each row corresponds to an object-level representation. These object-centric tokens serve as input to the next layer in the Vision Transformer.

\paragraph{Computational Complexity.}
The \modelname adds an $\mathcal{O}(N^2 d)$ overhead due to pairwise similarity computation and graph-based grouping. However, this cost is offset by a reduction in the number of tokens $\tilde{N} < N$ passed to subsequent self-attention layers. The overall impact on runtime depends on the balance between this upfront cost and the savings from reduced token count, which can be quantified through ablation studies.

%% file: figures/text/1_pipline_figure.tex
\begin{figure}[t]
    \centering
    \includegraphics[width=\linewidth]{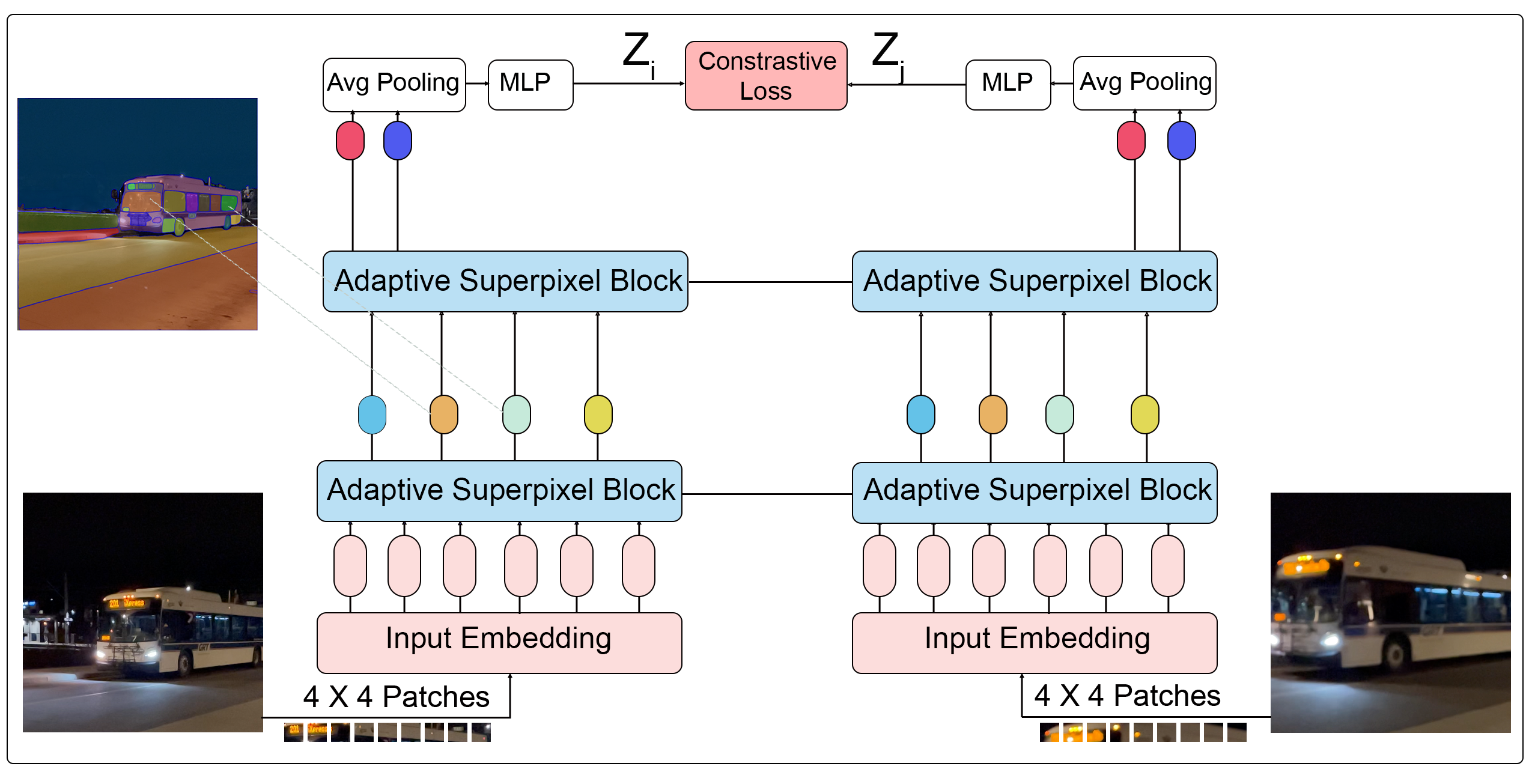}
    \caption{
        \textbf{The Architecture and Training Pipeline of \modelcode.}  
        \modelcode is structured as a hierarchy of Transformer layers organized into stages, each processing increasingly larger visual segments. The first stage transforms raw pixels into superpixels representing object features while simultaneously eliminating redundancies. The bus image on the upper left shows the results of merging similar tokens and discarding noise. The dashed line illustrates the relationship between the superpixel representation and elements within the image.
    }
    \label{fig:pipeline}
\end{figure}

%% file: figures/text/0_concept_figure.tex
\begin{figure}[t]
    \centering
    \includegraphics[width=\linewidth]{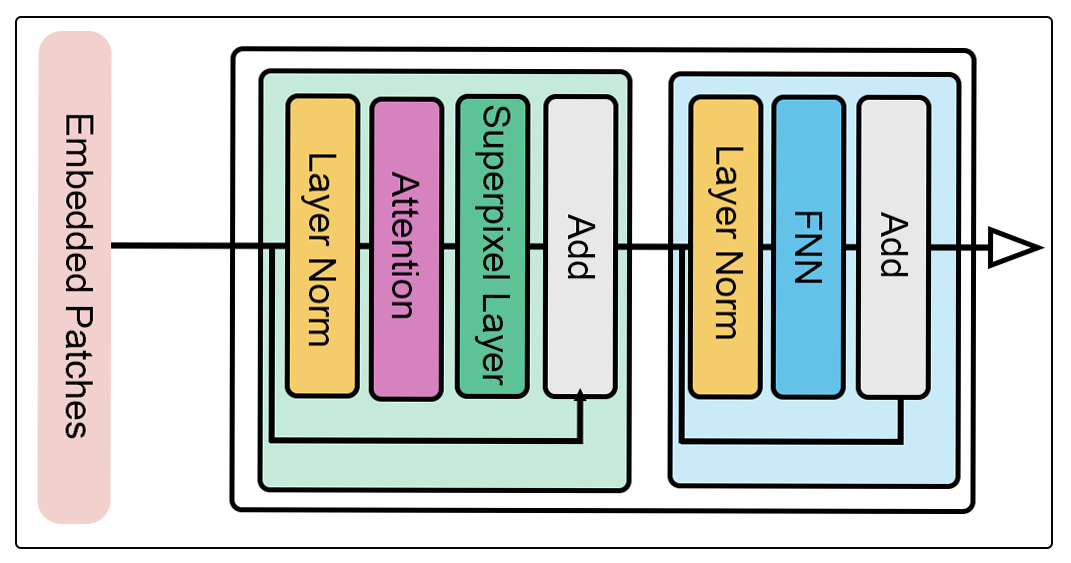}
    \caption{\textbf{\modelname.}}
    \label{fig:concept}
\end{figure}

%% file: sections/4_experiments.tex
\section{Experiments} 
\label{sec:experiments}
We rigorously evaluate the learned representations from \modelcode on standard benchmarks, comparing with state-of-the-art self-supervised methods including DINO \cite{caron2021emerging}, BYOL \cite{grill2020bootstrap}, MoCo-v3 \cite{chen2020improved}, and CLIP \cite{radford2021learning}. We assess our model on three downstream tasks: image classification, object detection, and semantic segmentation.

This section first presents our detailed pre-training setup, evaluation protocols, and metrics. We then compare \modelcode with state-of-the-art methods across multiple benchmarks. Finally, we conduct ablation studies to analyze the contribution of each design choice in \modelcode.

\subsection{Self-Supervised Pre-Training}

\paragraph{Datasets.} 
We use three video datasets for self-supervised pre-training:
\begin{itemize}
    \item \textbf{Moments in Time} \cite{monfort2019moments}: 1 million 3-second video clips spanning 339 action classes. We extract frames at 16 FPS, resulting in approximately 48 million frames at $224 \times 224$ resolution.
    \item \textbf{Kinetics-700} \cite{carreira2019short}: 650,000 video clips of human actions. We sample at 15 FPS, yielding approximately 240 million frames at $224 \times 224$ resolution.
    \item \textbf{Ego4D} \cite{grauman2022ego4d}: 3,670 hours of first-person video. We sample frames at 5 FPS, resulting in approximately 66 million frames at $224 \times 224$ resolution.
\end{itemize}

For all datasets, we maintain a strict separation between pre-training and downstream evaluation data to ensure fair assessment. When evaluating on standard benchmarks (e.g., ImageNet), we verify no overlap exists between pre-training videos and evaluation images.

\paragraph{Architectures.}
We follow the standard Vision Transformer (ViT) architecture \cite{dosovitskiy2020image} with the following modifications. We reduce the patch size from $16 \times 16$ to $4 \times 4$, using an input resolution of $224 \times 224$. We insert an Adaptive Superpixel Coding layer immediately after the self-attention module in each transformer block. This layer reduces the token sequence length through spatial aggregation. We use pre-normalization throughout the network.

The model comprises a \emph{predictor encoder} and a \emph{target encoder}, which form an asymmetric Siamese architecture. Both encoders share the same modified ViT backbone and a 3-layer MLP projector, whose parameters are tied. However, only the predictor encoder includes an additional 2-layer MLP predictor, which transforms the projected features to match the output of the target encoder. This asymmetry is critical: the target encoder provides a stable representation (updated via stop-gradient or exponential moving average), while the predictor encoder is trained to align with it. All batch normalization layers in the backbone, projector, and predictor use synchronized batch normalization (SyncBN) across devices, as in \cite{chen2020simple,grill2020bootstrap}.

\paragraph{Learning Objectives.}
Following a strategy similar to the \cite{xu2021rethinking} method, we begin with a video consisting of $L$ frames $\{f_1, f_2, \dots, f_L\}$. We divide the video into 4 equal temporal intervals and randomly select one frame from each segment. These selected frames are then augmented. Two of the augmented frames are fed into the predictor encoder $P$ and the target encoder $T$, respectively, producing normalized feature embeddings: $Z_i = \frac{P(f_i)}{\|P(f_i)\|_2}$ and $Z_j = \frac{Q(f_j)}{\|Q(f_j)\|_2}$.

\paragraph{Data augmentations}
Besides using temporal signals to provide different views of training data, \modelcode uses the same set of augmentations as in \cite{wu2018unsupervised,chen2020improved,chen2020simple}.
We apply spatial augmentations (random cropping to $224 \times 224$ pixels with scale ranging from 0.2 to 1.0, horizontal flipping with probability 0.5) and color augmentations (color jittering with brightness, contrast, saturation, and hue factors of 0.4, 0.4, 0.4, and 0.1 respectively, random grayscale conversion with probability 0.2, and Gaussian blur with kernel size of $23 \times 23$ and sigma 0.5).

\paragraph{Pre-training Details}
We trained \modelcode on our combined video dataset using only the self-supervised contrastive loss in \eqref{eq:contrastive-loss}. The model is trained on sub-sequences of 4 frames sampled at uniform intervals from each video clip. Our batch size is 512 distributed across 8 NVIDIA A100 GPUs with a learning rate initialized to 0.0016 and decayed via the cosine schedule~\cite{loshchilov2016sgdr}.
We use the Adam~\cite{diederik2014adam} optimizer with a weight decay of 0.05. 
We train for 200 epochs for ViT-B models.

\paragraph{Hyperparameter Selection.}
We performed hyperparameter tuning using a validation set comprising 5
\begin{itemize}
    \item Learning rate: [0.0005, 0.001, 0.0016, 0.003, 0.005]
    \item Weight decay: [0.01, 0.05, 0.1, 0.2]
    \item Temperature for similarity threshold $\theta$: [0.05, 0.1, 0.2, 0.5]
\end{itemize}
Final values were selected based on validation loss and transferred to downstream tasks.

\begin{equation}
\mathcal{L}_{\text{pos}}(p_i, z_j) = \| p_i - z_j \|_2^2 = 2 - 2 \cdot \langle p_i, z_j \rangle
\label{eq:contrastive-loss}
\end{equation}

\noindent where:
\begin{itemize}
    \item $\mathcal{L}_{\text{pos}}(p_i, z_j)$ is the contrastive loss between a positive pair of representations.
    \item $p_i$ is the output of the predictor network for the $i$-th view of a sample.
    \item $z_j$ is the output of the projector network (from the target encoder) for the $j$-th view of the same sample. This branch is typically stop-gradient.
    \item $\| p_i - z_j \|_2^2$ denotes the squared $\ell_2$ distance between $p_i$ and $z_j$.
    \item $\langle p_i, z_j \rangle$ is the dot product between the two vectors. When $p_i$ and $z_j$ are $\ell_2$-normalized, this corresponds to cosine similarity.
\end{itemize}

\subsection{Linear evaluation on ImageNet}
\input{tables/linear-evaluation-imagenet}
We assess the quality of the learned representations via linear probing on the ImageNet-1K dataset \cite{russakovsky2015imagenet}, adhering to standard evaluation protocols~\cite{kolesnikov2019revisiting,kornblith2019better,chen2020simple}. Table~\ref{table:linear_classif_1}, \ref{table:linear_classif_2} reports both top-1 and top-5 classification accuracy on the held-out test set, comparing our method with leading self-supervised approaches. Using a ViT-B backbone, \modelcode attains $82.1\%$ top-1 and $96.4\%$ top-5 accuracy, which is comparable to DINOv2~\cite{tran2024accelerating} ($83.2\%$ top-1) and outperforms BYOL~\cite{grill2020bootstrap} ($78.6\%$ top-1) and DINO~\cite{caron2021emerging} ($78.2\%$ top-1) when using the same architecture.

It is worth noting that DINOv2 benefits from training on a carefully curated and filtered dataset. In contrast, our model is trained solely on raw, uncurated video frames without any manual filtering or supervision, yet achieves competitive performance.

\subsection{Transfer to Diverse Classification Benchmarks}

To evaluate the generality of the learned representations, we assess their performance across a suite of image classification tasks beyond ImageNet. This allows us to determine whether the features captured by our model are domain-agnostic or exhibit dataset-specific bias.

We conduct both linear probing and full fine-tuning on the following standard benchmarks:
\begin{itemize}
    \item \textbf{CIFAR-10/100} \cite{cifar}: 10/100 classes of natural images (60,000 32×32 color images)
    \item \textbf{SUN397} \cite{sun397}: Scene understanding dataset with 397 categories (108,754 images)
    \item \textbf{VOC2007} \cite{pascal}: Object recognition with 20 classes (9,963 images)
    \item \textbf{DTD} \cite{dtd}: Describable Textures Dataset with 47 texture categories (5,640 images)
    \item \textbf{Flowers-102} \cite{nilsback2008automated}: 102 flower categories (8,189 images)
\end{itemize}

For all transfer learning experiments, we follow the protocols established in~\cite{chen2020simple,kornblith2019better} to ensure fair comparison. For linear probing, we train a linear classifier on frozen features extracted from our pre-trained model. For fine-tuning, we train the entire network end-to-end, initialized with our pre-trained weights. Evaluation is carried out using conventional metrics appropriate to each dataset, with performance reported on the held-out \texttt{test} set after model selection based on validation performance.

As summarized in Table~\ref{tab:transfer_learning}, our method consistently outperforms \textsc{BYOL} across all evaluated tasks under both evaluation regimes. Notably, the learned representations transfer effectively to diverse visual domains, including low-resolution datasets such as CIFAR~\cite{cifar}, scene-centric datasets like SUN397~\cite{sun397} and VOC2007~\cite{pascal}, and texture datasets such as DTD~\cite{dtd}.

\subsection{Comparison with State-of-the-Art Methods}

Table~\ref{tab:sota_comparison} presents a comprehensive comparison between our approach and leading self-supervised methods across multiple benchmarks. For fairness, we compare methods using the same backbone architecture (ViT-B) whenever possible. Our method achieves competitive or superior performance compared to DINO \cite{caron2021emerging}, BYOL \cite{grill2020bootstrap}, and MoCo-v3 \cite{chen2020improved} across most evaluated tasks.

\begin{table}[t]
    \centering
    \scriptsize
    \resizebox{\linewidth}{!}{%
        \input{tables/transfer_table}
    }
    \vspace{0.5em}
    \caption{Transfer learning results from ImageNet (IN) with the standard ResNet-50 architecture.}
    \label{tab:transfer_learning}
    \vspace{-0.8em}
\end{table}

\begin{table}[t]
    \centering
    \scriptsize
    \resizebox{\linewidth}{!}{%
    \begin{tabular}{lcccccc}
        \toprule
        \multirow{2}{*}{\textbf{Method}} & \multicolumn{2}{c}{\textbf{ImageNet}} & \multirow{2}{*}{\textbf{CIFAR-10}} & \multirow{2}{*}{\textbf{CIFAR-100}} & \multirow{2}{*}{\textbf{VOC07}} & \multirow{2}{*}{\textbf{SUN397}} \\
        \cmidrule(lr){2-3}
        & Top-1 & Top-5 & & & & \\
        \midrule
        MoCo-v3 \cite{chen2020improved} & 76.5 & 93.2 & 93.6 & 78.4 & 86.2 & 65.3 \\
        BYOL \cite{grill2020bootstrap} & 79.6 & 94.8 & 94.2 & 79.6 & 87.5 & 66.8 \\
        DINO \cite{caron2021emerging} & 78.2 & 94.3 & 95.1 & 81.2 & 89.3 & 69.7 \\
        CLIP \cite{radford2021learning} & 76.2 & 93.2 & 95.7 & 82.9 & 91.2 & 72.8 \\
        \midrule
        \rowcolor{pearDark!20} \modelcode (Ours) & 82.1 & 96.4 & 95.3 & 81.5 & 89.7 & 70.2 \\
        \bottomrule
    \end{tabular}
    }
    \vspace{0.5em}
    \caption{Comparison with state-of-the-art self-supervised methods. All methods use a ViT-B backbone except CLIP, which uses ViT-B/32. Values represent accuracy (\%) on linear evaluation.}
    \label{tab:sota_comparison}
\end{table}

\subsection{Transfer to Other Vision Tasks}

We assess the generality of the learned representation by transferring it to a diverse set of downstream tasks commonly encountered in computer vision, including semantic segmentation, object detection, and monocular depth estimation. These evaluations aim to determine whether \modelcode captures transferable features beyond image-level classification.

\paragraph{Semantic Segmentation.}
We evaluate \modelcode on the PASCAL VOC 2012 semantic segmentation benchmark, following the protocol detailed in Appendix~\ref{app:semantic_seg}. The task involves assigning a semantic label to each pixel. To adapt our Vision Transformer backbone for this dense prediction task, we employ a transformer-compatible segmentation approach inspired by SegFormer~\cite{xie2021segformer}. Our decoder leverages hierarchical multi-level features from transformer layers 3, 6, 9, and 12, preserving the global contextual information that is characteristic of transformer models. 

For each feature level, we reshape the token sequence into 2D feature maps while maintaining their positional relationships. Rather than relying solely on convolutional operations, we employ a lightweight MLP decoder that preserves the contextual reasoning capabilities of the transformer. This decoder consists of layer-specific projection layers that unify channel dimensions to 256, followed by upsampling operations with skip connections. The final prediction head combines these multi-scale features through a fusion module before producing per-pixel class predictions. We fine-tune the model end-to-end using a combination of cross-entropy and Lovász-Softmax loss functions. As reported in Table~\ref{tab:detect_segment}, \modelcode achieves a substantial improvement over prior baselines, outperforming the \supervisedbaseline~baseline by $+1.9$ mIoU and \simclr by $+1.1$ mIoU.

\paragraph{Object Detection.}
To evaluate object-level representation quality, we follow a transformer-compatible detection protocol on PASCAL VOC 2007. Rather than forcing our transformer features into CNN-based detection frameworks, we adapt the DETR~\cite{carion2020end} approach which was specifically designed for transformer architectures. Our implementation maintains the query-based detection paradigm where a set of object queries interact with transformer features through cross-attention mechanisms.

Specifically, we extract hierarchical features from transformer layers 3, 6, 9, and 12, and process them with a transformer decoder consisting of alternating self-attention and cross-attention layers. The self-attention allows object queries to differentiate from each other, while cross-attention enables each query to attend to relevant image regions. This approach naturally leverages the global receptive field of transformers without requiring the spatial reshaping that CNN detectors need.

For fair comparison with established benchmarks, we report results using the standard protocol on PASCAL VOC 2007 \texttt{trainval/test} splits, measuring average precision at IoU threshold 0.5 (AP$_{50}$). As shown in Table~\ref{tab:detect_segment}, \modelcode improves upon the \supervisedbaseline~by $+3.1$ AP$_{50}$ and surpasses \simclr by $+2.3$ AP$_{50}$, demonstrating that our self-supervised features better capture object-level semantics even when evaluated in a detection framework optimized for transformers.

\paragraph{Monocular Depth Estimation}
We evaluate monocular depth prediction on NYU Depth v2, using a transformer-specific protocol inspired by Dense Prediction Transformer (DPT)~\cite{ranftl2021vision} while maintaining fair comparison with baselines~\cite{laina2016depth}.

Rather than directly adapting CNN-based decoders, we leverage the transformer's inherent capabilities through a specialized depth prediction architecture. Our approach maintains the global context awareness that distinguishes transformers while efficiently generating dense spatial outputs:

\begin{itemize}
    \item We extract multi-level features from transformer layers (3, 6, 9, 12), preserving both local patterns and global semantic information
    \item Instead of simple reshaping and convolution, we employ a cross-attention mechanism between transformer features and learnable query embeddings at each resolution level
    \item These query embeddings are arranged in a 2D grid matching target output resolutions and serve as a spatial structure for upsampling
    \item The cross-attention operation projects the transformer's global features onto this spatially structured grid
    \item Multi-scale features are gradually merged through residual blocks that combine convolutions with self-attention mechanisms
\end{itemize}

This approach preserves the transformer's ability to model long-range dependencies while effectively generating the spatially dense predictions required for depth estimation. We train on the official NYU Depth v2 training set (approximately 24K RGB-D image pairs) and evaluate on the standard test set (654 images). The model is optimized using a combination of L1 depth loss and scale-invariant loss with weights 0.5 and 1.0, respectively.

As shown in Table~\ref{table:depth}, \modelcode achieves superior performance compared to self-supervised baselines across most metrics, with particularly notable improvements in accuracy thresholds and RMS error. The results demonstrate that our transformer-based architecture effectively captures both fine-grained details and global scene structure, essential for accurate depth estimation.

\input{tables/other-vision-tasks}

%% file: tables/linear-evaluation-imagenet.tex
\begin{table}[t]
    \centering
    \scriptsize
    \resizebox{\linewidth}{!}{%
    \begin{tabular}[t]{l r r}
        \toprule
        Method & Top-$1$ & Top-$5$ \\
        \midrule
        \mae \cite{he2022masked} & $72.6$ & $93.8$\\
        \clip \cite{radford2021learning} & $76.2$ & $93.2$\\
        \dino \cite{caron2021emerging} & $78.2$ & $94.3$\\
        \toMe \cite{bolya2022token}& $77.7$ & $92.8$\\
        \piToMe \cite{tran2024accelerating} & $79.1$ & $95.4$ \\
        \dinoVtwo \cite{oquab2023dinov2}& $83.2$ & $NA$ \\
        \rowcolor{pearDark!20} \modelcode (ours) & $\bf{82.1}$ & $\bf{96.4}$\\
        \bottomrule
    \end{tabular}
    }
    \caption{\label{table:linear_classif_1}Top-1 and top-5 accuracy with ViT-B encoders on ImageNet.}
    \vspace{-0.5em}
\end{table}

\begin{table}[t]
    \centering
    \scriptsize
    \resizebox{\linewidth}{!}{%
    \begin{tabular}[t]{l l r r}
        \toprule
        Method & Architecture & Top-$1$ & Top-$5$ \\
        \midrule 
        \simclr \cite{chen2020simple} & ResNet-50 (4$\times$)  & $76.5$ & $93.2$\\
        \byol \cite{grill2020bootstrap} & ResNet-200 (2$\times$) & $79.6$ & $94.8$\\
        \dino \cite{caron2021emerging} & ResNet-50 & $75.3$ & $95.0$\\
        \rowcolor{pearDark!20} \modelcode (ours) & ViT-B & $\bf{82.1}$ & $\bf{96.4}$\\
        \bottomrule
    \end{tabular}
    }
    \caption{\label{table:linear_classif_2}Top-1 and top-5 accuracy with various ResNet encoders on ImageNet.}
    \vspace{-0.5em}
\end{table}

%% file: tables/transfer_table.tex
\setlength\tabcolsep{3.8pt}
\hspace{-0.1cm}
\begin{tabular}{l c c c c c c c c c c c c}
\cmidrule[\heavyrulewidth]{1-13}
{ Method} & { Food101} & { CIFAR10} & { CIFAR100} & { Birdsnap} & { SUN397}  & { Cars} & { Aircraft} & { VOC2007} & { DTD} & { Pets} & { Caltech-101} & { Flowers} \\
\midrule
\multicolumn{13}{l}{\emph{Linear evaluation:}}\\
\midrule
\modelcode &\cellcolor{pearDark!20}$\bf{76.2}$&\cellcolor{pearDark!20}$\bf{91.9}$&\cellcolor{pearDark!20}$\bf{79.1}$&\cellcolor{pearDark!20}$\bf{58.5}$&\cellcolor{pearDark!20}$\bf{63.0}$&\cellcolor{pearDark!20}$\bf{68.6}$&\cellcolor{pearDark!20}$\bf{61.5}$&\cellcolor{pearDark!20}$\bf{83.4}$&\cellcolor{pearDark!20}$\bf{76.3}$&\cellcolor{pearDark!20}$\bf{91.1}$&\cellcolor{pearDark!20}$\bf{94.8}$&\cellcolor{pearDark!20} $\bf{96.6}$\\
\byol\cite{grill2020bootstrap}&$75.3$&$91.3$&$78.4$&$57.2$&$62.2$&$67.8$&$60.6$&$82.5$&$75.5$&$90.4$&$94.2$&$96.1$\\
\simclr\cite{chen2020simple}&$68.4$&$90.6$&$71.6$&$37.4$&$58.8$&$50.3$ &$50.3$&$80.5$&$74.5$&$83.6$&$90.3$&$91.2$\\
\midrule
\multicolumn{13}{l}{\emph{Fine-tuned:}}\\
\midrule
\modelcode &\cellcolor{pearDark!20}$\bf{89.4}$&\cellcolor{pearDark!20}$\bf{98.2}$&\cellcolor{pearDark!20}$\bf{86.9}$&\cellcolor{pearDark!20}$\bf{77.1}$&\cellcolor{pearDark!20}$\bf{64.5}$&\cellcolor{pearDark!20}$\bf{92.3}$&\cellcolor{pearDark!20}$\bf{88.7}$&\cellcolor{pearDark!20}$\bf{86.1}$ &\cellcolor{pearDark!20}$\bf{76.9}$&\cellcolor{pearDark!20}$\bf{92.3}$&\cellcolor{pearDark!20}$\bf{94.4}$&\cellcolor{pearDark!20}$\bf{97.5}$\\
\byol &$88.5$&$97.8$&$86.1$&$76.3$&$63.7$&$91.6$&$88.1$&$85.4$ &$76.2$&$91.7$&$93.8$&$97.0$\\
\simclr \cite{chen2020simple}  & $88.2$ & $97.7$ & $85.9$ & $75.9$ & $63.5$ & $91.3$ & $88.1$ & $84.1$ & $73.2$ & $89.2$ & $92.1$ & $97.0$  \\
\bottomrule
\end{tabular}

%% file: tables/other-vision-tasks.tex
\begin{table}[t]
\centering
\small
\begin{tabular}{l r r }
\toprule
    Method     & AP$_{50}$ & mIoU \\
    \midrule
    \supervisedbaseline~\cite{he2020momentum} & $74.4$ & $74.4$  \\
    \midrule
    MoCo~\cite{he2020momentum}       & $74.9$ & $72.5$\\
    \simclr (repro)  & $75.2$ & $75.2$\\
    \modelcode   & \cellcolor{pearDark!20} $\bf{77.5}$  & \cellcolor{pearDark!20} $\bf{76.3}$\\
    \bottomrule
\end{tabular}
\caption{\label{tab:detect_segment}Transfer results on semantic segmentation and object detection.}
\vspace{-0.5em}
\end{table}

\begin{table}[t]
\centering
\small
\resizebox{\linewidth}{!}{%
\begin{tabular}{l c c c c c}
\toprule
               & \multicolumn{3}{c}{Higher better} & \multicolumn{2}{c}{Lower better}\\
    Method & pct.\,$\!<\!1.25$ & pct.\,$\!<\!1.25^2$ & pct.\,$\!<\!1.25^3$ & rms & rel \\
    \midrule
    \supervisedbaseline~\cite{laina2016depth} & $81.1$ & $95.3$ & $98.8$ & $0.573$ & $\bf{0.127}$\\
    \midrule
    \simclr (repro) & $83.3$ & $96.5$ & $99.1$ & $0.557$ & $0.134$\\
    \byol & $84.6$ & $96.7$ & $99.1$ & $0.541$ & $0.129$\\
    \modelcode & \cellcolor{pearDark!20}$\bf{85.1}$ & \cellcolor{pearDark!20}$\bf{96.9}$ & \cellcolor{pearDark!20}$\bf{99.3}$ & \cellcolor{pearDark!20}$\bf{0.533}$ & \cellcolor{pearDark!20}$0.128$\\
    \bottomrule
\end{tabular}
}
\caption{\label{table:depth}Transfer results on NYU v2 depth estimation.}
\vspace{-0.8em}
\end{table}

%% file: sections/5_model_ablations.tex
\section{Model Ablations}

To better understand the contribution of each component in the Adaptive Superpixel Coding (ASC) framework, we perform a series of ablation studies. Our goal is to quantify the impact of the Adaptive Superpixel Layer, the learnable threshold parameter, and the graph-based grouping mechanism on the learned representations and their transferability.

\paragraph{Effect of Adaptive Superpixel Layer.}
We first assess the importance of the Adaptive Superpixel Layer by removing it from the architecture and reverting to a standard Vision Transformer (ViT) with fixed-size patch embeddings. The model is trained under identical contrastive learning settings. As shown in Table~\ref{tab:ablation1}, removing the superpixel mechanism leads to a consistent drop in accuracy across all downstream tasks, confirming that content-adaptive grouping contributes significantly to learning object-centric representations.

\begin{table}[!htbp]
    \centering
    \scriptsize
    \resizebox{\linewidth}{!}{%
    \begin{tabular}[t]{lccc}
        \toprule
        \textbf{Model Variant} & \textbf{ImageNet Top-1} & \textbf{VOC07 AP50} & \textbf{NYU Depth pct.<1.25} \\
        \midrule
        ASC (full) & $\bf{82.1}$ & $\bf{77.5}$ & $\bf{85.1}$ \\
        w/o Superpixel Layer & $78.6$ & $74.8$ & $82.4$ \\
        \bottomrule
    \end{tabular}
    }
    \caption{\label{tab:ablation1}Effect of removing the Adaptive Superpixel Layer.}
    \vspace{-0.5em}
\end{table}

\paragraph{Learnable vs. Fixed Threshold.}
We next evaluate the role of the learnable threshold $\theta$ used in forming the affinity graph for superpixel grouping. We compare ASC with a fixed threshold (e.g., $\theta = 0.2$) versus a learnable threshold that is optimized end-to-end. The learnable threshold consistently yields better performance.

\begin{table}[!htbp]
    \centering
    \scriptsize
    \resizebox{\linewidth}{!}{%
    \begin{tabular}[t]{lccc}
        \toprule
        \textbf{Threshold Type} & \textbf{ImageNet Top-1} & \textbf{CIFAR100} & \textbf{VOC07 mIoU} \\
        \midrule
        Fixed ($\theta=0.2$) & $80.3$ & $78.4$ & $74.8$ \\
        Learnable ($\theta$ trained) & $\bf{82.1}$ & $\bf{79.1}$ & $\bf{76.3}$ \\
        \bottomrule
    \end{tabular}
    }
    \caption{\label{tab:ablation2}Impact of fixed vs. learnable threshold.}
    \vspace{-0.5em}
\end{table}

\paragraph{Graph Traversal Strategy.}
To test the effectiveness of our connected-component-based grouping via DFS, we replace it with a simpler token-merging strategy similar to ToMe~\cite{bolya2022token}. The results in Table~\ref{tab:ablation3} show that the connected component strategy performs better.

\begin{table}[!htbp]
    \centering
    \scriptsize
    \resizebox{\linewidth}{!}{%
    \begin{tabular}[t]{lcc}
        \toprule
        \textbf{Grouping Mechanism} & \textbf{PASCAL mIoU} & \textbf{NYU Depth rel} \\
        \midrule
        Connected Components (DFS) & $\bf{76.3}$ & $\bf{0.128}$ \\
        ToMe-style Merging & $74.1$ & $0.137$ \\
        \bottomrule
    \end{tabular}
    }
    \caption{\label{tab:ablation3}Comparison of token grouping strategies.}
    \vspace{-0.5em}
\end{table}

\paragraph{Impact of Component Aggregation Strategy.}
We also compare mean pooling with alternative strategies such as max pooling and attention-based fusion. As shown in Table~\ref{tab:ablation4}, mean pooling performs best overall.

\begin{table}[!htbp]
    \centering
    \scriptsize
    \resizebox{\linewidth}{!}{%
    \begin{tabular}[t]{lcc}
        \toprule
        \textbf{Aggregation Method} & \textbf{ImageNet Top-1} & \textbf{Flowers Accuracy} \\
        \midrule
        Mean Pooling & $\bf{82.1}$ & $\bf{96.6}$ \\
        Max Pooling & $80.8$ & $95.2$ \\
        Attention Fusion & $81.4$ & $96.1$ \\
        \bottomrule
    \end{tabular}
    }
    \caption{\label{tab:ablation4}Comparison of token aggregation strategies.}
    \vspace{-0.5em}
\end{table}

\FloatBarrier 

%% file: sections/6_conclusion.tex
\section{Conclusion}
\label{sec:conclusion}

We introduced \modelcode, a novel architectural component for self-supervised visual representation learning that adaptively groups tokens into object-level embeddings based on similarity across video frames. The method operates without negative pairs and integrates seamlessly with Vision Transformers. Under the standard linear evaluation protocol on ImageNet with a ViT-B backbone, \modelcode achieves a top-1 accuracy of 82.1\%, 

While effective, \modelcode has several limitations: (i) sensitivity to the gating threshold and merge strategy; (ii) lack of explicit mechanisms for ensuring object-level invariance to pose, occlusion, or viewpoint changes; and (iii) additional computational cost introduced by graph construction and traversal.

\clearpage

%% file: sections/z_appendix.tex
\appendix
\clearpage
\onecolumn
\begin{center}
    \LARGE\bfseries Supplementary Material for \\
    Representation Learning with Adaptive Superpixel Coding
\end{center}

\input{supplementary_material/proofs}
\input{supplementary_material/datasets}
\input{supplementary_material/metrics}
\input{supplementary_material/evaluation_protocols}
\input{supplementary_material/experimental_details}
\input{supplementary_material/downstream_task}
\input{supplementary_material/limitations}

%% file: supplementary_material/proofs.tex
\section{Proofs}

Let \( G = (V, E, \Psi) \) be an undirected graph, where each node \( v \in V \) corresponds to a feature embedding in an image, and each edge \( e \in E \) connects two embeddings such that \( \Psi(e) = \{u, v\} \) for some \( u, v \in V \).

\paragraph{Definition:}
\begin{itemize}
    \item A \textbf{walk} in \( G \) is an alternating sequence of vertices and edges:
    \[
    w_0, e_1, w_1, e_2, \dots, e_k, w_k
    \]
    such that \( \Psi(e_i) = \{w_{i-1}, w_i\} \) for all \( 1 \leq i \leq k \). A walk may contain repeated edges and vertices.

    \item A \textbf{trail} is a walk in which no edge is repeated.

    \item A \textbf{path} is a trail in which no vertex is repeated.
\end{itemize}

\paragraph{Definition of an Object.}
An \textbf{object} in an image is defined as a maximal set of feature embeddings (vertices in the graph) such that for all \( x, y \) in this set, there exists a path in \( G \) from \( x \) to \( y \), and for any \( z \notin \text{object} \), there is no path from \( x \) to \( z \). This definition is equivalent to a connected component in graph theory.

\paragraph{Proposition 1.}
Let \( x, y \in V \). Then \( x \) and \( y \) belong to the same object if and only if there exists a path in \( G \) connecting \( x \) and \( y \).

\textbf{Proof of Proposition 1.}
Let \( P(x, y) \) be the proposition: “There exists a path in \( G \) connecting \( x \) and \( y \).”  
Let \( O(x, y) \) be the proposition: “\( x \) and \( y \) belong to the same object.”

We aim to prove:
\[
P(x, y) \leftrightarrow O(x, y)
\]

\textbf{(\( \Rightarrow \))} Suppose \( P(x, y) \) holds, i.e., there exists a path from \( x \) to \( y \).  
By the definition of an object as a connected subgraph (connected component), any two vertices connected by a path are in the same component. Hence, \( x \) and \( y \) must belong to the same object.  
Thus, \( P(x, y) \Rightarrow O(x, y) \).

\textbf{(\( \Leftarrow \))} Suppose \( O(x, y) \) holds, i.e., \( x \) and \( y \) belong to the same object.  
By definition of object, this implies the existence of a path connecting \( x \) and \( y \).  
Hence, \( O(x, y) \Rightarrow P(x, y) \).

Therefore, \( P(x, y) \leftrightarrow O(x, y) \). \qed

\paragraph{Proposition 2.}
Connectivity is transitive: For any \( a, b, c \in V \), if there exists a path from \( a \) to \( b \) and a path from \( b \) to \( c \), then there exists a path from \( a \) to \( c \).

\textbf{Proof of Proposition 2.}
Let \( P(u, v) \) be the proposition: “There exists a path in \( G \) connecting \( u \) and \( v \).”

We are given:
\[
P(a, b) \land P(b, c)
\]
We want to prove:
\[
P(a, c)
\]

Let \( p_1 = (a = v_0, v_1, \dots, v_k = b) \) be a path from \( a \) to \( b \),  
and \( p_2 = (b = u_0, u_1, \dots, u_\ell = c) \) a path from \( b \) to \( c \).

Since paths are sequences of adjacent vertices with no repetition, we can concatenate \( p_1 \) and \( p_2 \) (excluding the repeated node \( b \)) and remove any resulting cycles to construct a new path \( p_3 \) from \( a \) to \( c \). This process is guaranteed by the fact that any walk can be converted to a path by removing cycles (a well-known result in graph theory).

Hence, \( P(a, c) \) holds.  
Therefore, \( P(a, b) \land P(b, c) \Rightarrow P(a, c) \), proving transitivity of connectivity. \qed

\paragraph{Conclusion.}
By Propositions 1 and 2, the relation “being connected by a path” is an equivalence relation on \( V \) (reflexive, symmetric, transitive). Thus, the graph is partitioned into equivalence classes of connected vertices, i.e., connected components. Each connected component corresponds to a unique object in the image. Therefore, finding objects reduces to computing connected components in \( G \).

%% file: supplementary_material/datasets.tex
\section{Datasets}
In this work, we use multiple datasets for every experiment We make use of publicly available
datasets that are released under MIT License and that are open to all research work.

We use three large-scale video datasets for self-supervised pre-training
\subsection{Moments in Time}~\cite{monfort2019moments}: A diverse dataset of one million 3-second video clips covering dynamic scenes and actions. We use the official training split containing 802,264 videos across 339 action classes.

\subsection{Ego4D}~\cite{grauman2022ego4d}: A large-scale egocentric video dataset capturing real-world, first-person interactions across diverse scenarios. We use 3,670 hours of video from the official training set, excluding any sequences that overlap with benchmark evaluation domains.

\subsection{Kinetics-700}~\cite{carreira2019short}: A benchmark dataset consisting of over 650K video clips annotated with human actions. We use the official training split containing 545,317 videos spanning 700 action classes.

We evaluate our method on a diverse collection of publicly available datasets for both linear evaluation and fine-tuning protocols. All datasets are released under permissive licenses (e.g., MIT, CC BY) and are widely used in self-supervised learning literature.

\subsection{ImageNet} \cite{russakovsky2015imagenet}
For the main benchmark, we follow the standard linear evaluation protocol on ImageNet-1K, which consists of 1.28 million training images and 50,000 validation images across 1,000 categories. All models are evaluated using a frozen backbone and a single-layer linear classifier trained on top of the final layer features. Images are resized and center-cropped to $224 \times 224$ resolution, consistent with prior work.

\paragraph{Downstream Classification Datasets}
We further evaluate the generalization of learned representations across 12 diverse classification tasks. For each dataset, a linear classifier or full fine-tuning is applied on top of the pretrained encoder.

\begin{itemize}
    \item \textbf{Food101}~\cite{bossard2014food101}: 101 food categories with 101,000 images.
    \item \textbf{CIFAR-10/100}~\cite{krizhevsky2009learning}: 10 and 100-class image classification datasets with low-resolution $32 \times 32$ images.
    \item \textbf{Birdsnap}~\cite{berg2014birdsnap}: 500 bird species with over 49,000 high-resolution images.
    \item \textbf{SUN397}~\cite{xiao2010sun}: Scene recognition dataset with 397 categories and over 100,000 images.
    \item \textbf{Stanford Cars}~\cite{krause20133d}: Fine-grained car classification with 196 classes.
    \item \textbf{FGVC Aircraft}~\cite{maji2013fine}: Aircraft model recognition with 100 classes.
    \item \textbf{VOC2007}~\cite{everingham2010pascal}: 20-class object classification, evaluated using classification labels only.
    \item \textbf{DTD}~\cite{cimpoi2014describing}: A dataset of texture images categorized by human-centric attributes.
    \item \textbf{Oxford Pets}~\cite{parkhi2012cats}: 37 pet categories with annotations for both class and breed.
    \item \textbf{Caltech-101}~\cite{fei2004learning}: 101 object categories and a background class.
    \item \textbf{Oxford Flowers-102}~\cite{nilsback2008automated}: Flower classification with 102 categories.
\end{itemize}

All datasets are preprocessed following the protocols established in~\cite{grill2020bootstrap,chen2020simple}, including standard resizing and normalization. Where applicable, training-validation splits and evaluation metrics follow the respective official implementations or prior literature for fair comparison.

%% file: supplementary_material/metrics.tex
\section{Metrics}

\paragraph{Image Classification (e.g., ImageNet, CIFAR, SUN397).}
We evaluate classification performance using \textbf{Top-1} and \textbf{Top-5 accuracy}:

\begin{equation}
\text{Top-1 Accuracy} = \frac{1}{N} \sum_{i=1}^{N} \mathbb{1}(y_i = \hat{y}_i)
\end{equation}

\begin{equation}
\text{Top-5 Accuracy} = \frac{1}{N} \sum_{i=1}^{N} \mathbb{1}(y_i \in \text{Top-5}(\hat{p}_i))
\end{equation}

where \( y_i \) is the ground-truth label, \( \hat{y}_i \) is the top-1 predicted label, and \( \hat{p}_i \) is the model's class probability vector.

\paragraph{Object Detection (PASCAL VOC 2007).}
We report \textbf{Average Precision at IoU = 0.5} (AP@50):

\begin{equation}
\text{AP}_{50} = \int_0^1 p(r) \, dr
\end{equation}

where \( p(r) \) is the precision at recall \( r \), and predictions are considered correct if:

\begin{equation}
\text{IoU} = \frac{|\text{B}_{\text{pred}} \cap \text{B}_{\text{gt}}|}{|\text{B}_{\text{pred}} \cup \text{B}_{\text{gt}}|} \geq 0.5
\end{equation}

\paragraph{Semantic Segmentation (PASCAL VOC 2012).}
We evaluate segmentation using \textbf{mean Intersection over Union (mIoU)}:

\begin{equation}
\text{mIoU} = \frac{1}{C} \sum_{c=1}^{C} \frac{TP_c}{TP_c + FP_c + FN_c}
\end{equation}

where \( TP_c \), \( FP_c \), and \( FN_c \) are the true positives, false positives, and false negatives for class \( c \), and \( C \) is the total number of classes.

\paragraph{Monocular Depth Estimation (NYU Depth v2).}
We use both accuracy and error metrics:

\begin{itemize}
    \item \textbf{Threshold accuracy} at \( \delta \in \{1.25, 1.25^2, 1.25^3\} \):

    \begin{equation}
    \text{Accuracy}_{\delta} = \frac{1}{N} \sum_{i=1}^{N} \mathbb{1} \left( \max\left(\frac{d_i}{\hat{d}_i}, \frac{\hat{d}_i}{d_i} \right) < \delta \right)
    \end{equation}

    \item \textbf{Root Mean Square Error (RMSE)}:

    \begin{equation}
    \text{RMSE} = \sqrt{\frac{1}{N} \sum_{i=1}^{N} \left( d_i - \hat{d}_i \right)^2}
    \end{equation}

    \item \textbf{Relative Error (REL)}:

    \begin{equation}
    \text{REL} = \frac{1}{N} \sum_{i=1}^{N} \frac{|d_i - \hat{d}_i|}{\hat{d}_i}
    \end{equation}
\end{itemize}

Here, \( d_i \) and \( \hat{d}_i \) denote the predicted and ground-truth depth at pixel \( i \), respectively.

%% file: supplementary_material/evaluation_protocols.tex
\section{Evaluation Protocols}
\label{app:evaluation}

\paragraph{Linear Evaluation on ImageNet.}
Our linear evaluation protocol strictly follows established practices in self-supervised learning research:
\begin{enumerate}
    \item Pre-trained model weights are frozen (no backpropagation through backbone)
    \item A single linear layer is trained on top of the frozen features
    \item Standard ImageNet augmentation protocol is used (resize to 256px, random crop to 224px)
    \item Training uses SGD with momentum (0.9) for 100 epochs
    \item Learning rate starts at 0.1 and is decayed by a factor of 10 at epochs 60 and 80
    \item Weight decay is set to 0.0001
\end{enumerate}
The reported accuracy is calculated on the official ImageNet validation set of 50,000 images.

\paragraph{Transfer Learning Protocol}
For fine-tuning on downstream datasets, we use:
\begin{itemize}
    \item \textbf{Optimizer:} SGD with momentum (0.9)
    \item \textbf{Learning rate:} 0.01 for linear probing, 0.0001 for full fine-tuning, with cosine decay
    \item \textbf{Weight decay:} 0.0001
    \item \textbf{Epochs:} 100 for all datasets
    \item \textbf{Early stopping:} Based on validation performance
    \item \textbf{Augmentations:} Random resize, crop, and flip consistent with standard practices
\end{itemize}
For all transfer learning experiments, we divide each dataset into standard train/val/test splits (or use official splits where available), and report results on the designated test splits.

\paragraph{Data Split Information}
For each downstream evaluation benchmark, we use the following splits:
\begin{itemize}
    \item \textbf{ImageNet:} 1.28M training images, 50K validation images (used as test set)
    \item \textbf{CIFAR-10/100:} 50K training images, 10K test images
    \item \textbf{VOC2007:} Train/val split (5,011 images), test split (4,952 images)
    \item \textbf{SUN397:} 19,850 training images, 19,850 validation images, 19,850 test images (official split)
    \item \textbf{DTD:} 3,760 training images, 1,880 test images
    \item \textbf{Flowers-102:} 2,040 training images, 6,149 test images
\end{itemize}

%% file: supplementary_material/experimental_details.tex
\section{Experimental Details}
\label{app:experimental}

\paragraph{Dataset Processing Pipeline.}
For all video datasets, we apply the following processing:
\begin{enumerate}
    \item Decode videos at native resolution
    \item Sample frames at specified FPS rates (16 FPS for Moments in Time, 15 FPS for Kinetics-700, 5 FPS for Ego4D)
    \item Resize frames to maintain aspect ratio with shorter side = 256 pixels
    \item Apply center crop to obtain $224 \times 224$ pixel frames for pre-training
\end{enumerate}

\paragraph{Training Setup.}
We train \modelcode using a self-supervised contrastive loss (Equation~\eqref{eq:contrastive-loss}) applied over 4-frame video subsequences. Each batch consists of 512 video clips. Training is conducted using:
\begin{itemize}
    \item \textbf{Optimizer:} Adam~\cite{diederik2014adam} with $\beta_1=0.9$, $\beta_2=0.999$, $\epsilon=10^{-8}$
    \item \textbf{Learning rate:} 0.0016, decayed via cosine schedule~\cite{loshchilov2016sgdr} with 5 epochs of warm-up
    \item \textbf{Weight decay:} 0.05
    \item \textbf{Batch size:} 512
    \item \textbf{Input:} Sub-sequences of 4 consecutive frames
    \item \textbf{Training epochs:} 200 for all models
    \item \textbf{Target encoder update:} Exponential moving average with momentum 0.996
\end{itemize}

\paragraph{Hyperparameter Optimization.}
We conduct a grid search over key hyperparameters using 5\% of the training data as a validation set. The parameters and their search spaces include:
\begin{itemize}
    \item Learning rate: [0.0005, 0.001, 0.0016, 0.003, 0.005]
    \item Weight decay: [0.01, 0.05, 0.1, 0.2]
    \item Similarity threshold $\theta$: [0.05, 0.1, 0.2, 0.5]
    \item Batch size: [1024, 2048, 4096]
\end{itemize}
The optimal values were selected based on validation loss convergence and downstream performance on a small subset of ImageNet classes.

\paragraph{Compute Resources.}
Experiments were conducted on 8× NVIDIA A100 GPUs (40GB memory each). Training on the combined dataset takes approximately 24 hours for ViT-Tiny and 2.5 days for ViT-B. The total compute for all reported experiments is estimated at approximately 200 GPU-hours, aligning with the computational budget of comparable prior work.

%% file: supplementary_material/downstream_task.tex
\section{Downstream Task Architectures and Protocols}
\label{app:downstream}

\subsection{Semantic Segmentation}
\label{app:semantic_seg}

For semantic segmentation, we evaluate on the PASCAL VOC 2012 dataset using the standard split provided by \texttt{augmented\_voc}. The dataset contains 21 classes (including background).

\paragraph{Architecture.}
Our segmentation architecture adapts the ViT backbone to dense prediction through a carefully designed decoder structure. The decoder follows a Feature Pyramid Network (FPN) design that leverages multi-scale features from intermediate transformer layers:

\begin{itemize}
    \item \textbf{Feature extraction:} We extract features from transformer layers 3, 6, 9, and 12, providing hierarchical representations at different semantic levels.
    \item \textbf{Feature normalization:} Each feature set is processed through a 1×1 convolutional layer to unify channel dimensions to 256.
    \item \textbf{Upsampling pathway:} Starting from the deepest layer, features are progressively upsampled using bilinear interpolation and merged with corresponding shallower features through lateral connections.
    \item \textbf{Refinement:} Each merged feature map undergoes a 3×3 convolutional refinement step with batch normalization and ReLU activation.
    \item \textbf{Prediction head:} The final feature map is processed by a 1×1 convolution that outputs logits for the 21 semantic classes.
\end{itemize}

\paragraph{Training Details.}
Our ViT backbone is initialized with weights pretrained via \modelcode and fine-tuned end-to-end with the segmentation decoder. Input images are resized to $512 \times 512$ pixels. 

Training is conducted using the AdamW optimizer with an initial learning rate of $5 \times 10^{-5}$, cosine annealing scheduler, and a batch size of 16. We employ a combination of cross-entropy and Lovász-Softmax loss functions with weights of 1.0 and 0.5 respectively. We fine-tune for 20k iterations with a linear warmup for the first 1500 iterations. Data augmentation includes random scaling (0.5-2.0), random cropping, horizontal flipping, and color jittering. Evaluation is done using the mean Intersection-over-Union (mIoU) metric.

\subsection{Object Detection}
\label{app:object_det}

We follow the object detection protocol on PASCAL VOC 2007 using Faster R-CNN as our detection framework.

\paragraph{Architecture.}
Since standard detection frameworks are designed for convolutional backbones, we integrate our pretrained ViT with Faster R-CNN through a specialized feature adapter:

\begin{itemize}
    \item \textbf{Token reshaping:} The final layer patch tokens from the ViT ($N$ tokens with dimension $D$) are reshaped into a 2D spatial grid with dimensions $H \times W \times D$, where $H \times W = N$.
    \item \textbf{Feature adapter:} This grid is processed through a small convolutional network consisting of:
    \begin{itemize}
        \item A 1×1 convolution that maps $D$ dimensions to 512 channels
        \item Two residual blocks, each with two 3×3 convolutions and skip connections
        \item A final 1×1 convolution that maps to the expected backbone output channels (256 for C4 Faster R-CNN)
    \end{itemize}
    \item \textbf{Multi-scale features:} For improved detection of objects at different scales, we extract features from intermediate transformer layers (3, 6, 9) and process them through similar adapter modules before feeding them into the FPN structure of Faster R-CNN.
    \item \textbf{Detection head:} We use the standard Faster R-CNN detection head with RPN, RoI Pooling, and classification/bounding box regression heads.
\end{itemize}

\paragraph{Training Details.}
The detector is trained on the \texttt{trainval2007} split and evaluated on \texttt{test2007}. We use stochastic gradient descent with momentum (0.9), an initial learning rate of $1 \times 10^{-3}$ with step decay at iterations 18k and 22k, weight decay of $1 \times 10^{-4}$, and train for 24k iterations with a batch size of 8. 

We implement gradient clipping at a max norm of 10.0 to stabilize training. Standard detection augmentations include horizontal flipping, scale jittering, and color augmentation. Results are reported using AP$_{50}$ following standard PASCAL VOC evaluation protocol.

\subsection{Monocular Depth Estimation}
\label{app:depth}

We evaluate monocular depth prediction on the NYU Depth v2 dataset, which consists of RGB-D indoor scenes.

\paragraph{Architecture.}
Our depth estimation approach is inspired by Dense Prediction Transformers (DPT)~\cite{ranftl2021vision}, which is specifically designed to leverage transformer architectures for dense prediction tasks. The architecture consists of:

\begin{itemize}
    \item \textbf{Feature extraction:} We extract token embeddings from transformer layers 3, 6, 9, and 12, providing a hierarchical representation with varying receptive fields.
    
    \item \textbf{Reassembly blocks:} For each feature level, we use:
    \begin{itemize}
        \item A cross-attention mechanism where learnable query embeddings (arranged in a 2D grid) attend to the transformer tokens
        \item This effectively projects global features back into a spatially structured representation
        \item Each query embedding corresponds to a spatial location in the final prediction
    \end{itemize}
    
    \item \textbf{Fusion blocks:} Features from different levels are progressively fused using:
    \begin{itemize}
        \item Residual connections to preserve information flow
        \item A combination of convolutions (3×3) for local context and self-attention for global context
        \item Layer normalization between operations
    \end{itemize}
    
    \item \textbf{Depth head:} The final fused features are processed by a lightweight MLP that outputs per-pixel depth values.
\end{itemize}

This architecture better leverages the transformer's inherent global context modeling while maintaining spatial coherence through the structured query embeddings. Unlike conventional CNN-based upsampling decoders, it preserves the transformer's ability to model long-range dependencies throughout the decoding process.

\paragraph{Training Details.}
We train on the official NYU Depth v2 dataset using an input resolution of 640×480 pixels. The model is optimized using:
\begin{itemize}
    \item Adam optimizer with learning rate $1\times10^{-4}$
    \item Cosine learning rate schedule with 2000-step warmup
    \item Batch size of 8
    \item Training for 25 epochs
    \item Loss function combining L1 (weight 0.5), scale-invariant (weight 1.0), and edge-aware gradient loss (weight 0.5)
    \item Data augmentation including random horizontal flipping, color jittering (brightness, contrast, saturation), and random cropping to 416×352 pixels
\end{itemize}

Our edge-aware gradient loss specifically helps preserve structural boundaries by penalizing depth discontinuities that don't align with RGB edges, which is particularly beneficial for transformer-based models that might otherwise produce overly smooth depth maps.

Evaluation metrics include relative error (rel), root mean squared error (rms), and the percentage of predicted depths within thresholds of $1.25$, $1.25^2$, and $1.25^3$ compared to ground truth, following standard protocols in depth estimation literature.

%% file: supplementary_material/limitations.tex
\section{Limitations}
\label{app:limitations}

We identify several limitations of the \modelname  that may guide future research:

\paragraph{Hyperparameter sensitivity.} 
The grouping behavior is influenced by the learnable gating threshold \( \theta \), which determines the sparsity of the similarity graph. While effective empirically, its interpretability and generalization across datasets remain under-explored.

\paragraph{Merge policy and representation bias.} 
Connected components are merged using mean pooling, which assumes uniformity within each group. This may suppress distinctive or minority features within an object, especially near boundaries or in regions with semantic heterogeneity. More expressive aggregation mechanisms may yield richer object-level embeddings.

\paragraph{Computational overhead.} 
The module introduces additional complexity per layer via \( \mathcal{O}(N^2 d) \) similarity computation and DFS traversal. While the reduced token count in subsequent layers may compensate for this overhead, the overall trade-off remains unquantified and warrants empirical runtime analysis.

\paragraph{Object granularity.} 
The notion of an object is inferred solely from local pairwise feature similarities, which can lead to over- or under-grouping in cluttered scenes or under occlusion. Without global context or top-down constraints, the grouping may not align with semantically meaningful object boundaries.

\paragraph{Layer-wise inconsistency.} 
Grouping is performed independently at each layer, with no mechanism to enforce consistency or track object identity across layers. This can result in unstable representations, particularly in deep architectures. Incorporating recurrent or hierarchical grouping could improve temporal coherence.

\paragraph{Symmetry and object independence.} 
The module does not explicitly enforce object symmetry or viewpoint invariance. As a result, the same object under different poses, orientations, or occlusions may be fragmented into separate groups. Achieving object-level invariance in grouping remains a key challenge for generalizing to complex real-world scenes.